\documentclass[letterpaper,journal]{IEEEtran}
\usepackage{amsmath,amsfonts,amssymb}
\usepackage{array}
\usepackage{textcomp}
\usepackage{stfloats}
\usepackage{url}
\usepackage{graphicx}
\usepackage{cite}
\usepackage{booktabs}
\usepackage[dvipsnames]{xcolor}
\definecolor{tbdred}{rgb}{0.85,0,0}\definecolor{TBDRED}{rgb}{0.85,0,0}

\newcommand{\Kb}{K_P^{\mathrm{B}}}

\usepackage[hidelinks]{hyperref}
\makeatletter
\def\ps@IEEEtitlepagestyle{%
  \def\@oddhead{\hbox{}\@IEEEheaderstyle\leftmark\hfil\thepage}\relax
  \def\@evenhead{\@IEEEheaderstyle\thepage\hfil\leftmark\hbox{}}\relax
  \def\@oddfoot{\hbox{}\hss\raisebox{\dimexpr\footskip-\preprintnoticedrop\relax}[0pt][0pt]{\parbox[t]{\textwidth}{\centering
     \normalfont\footnotesize\bfseries\color{red}This work has been submitted to the IEEE for possible publication. Copyright may be transferred without notice, after which this version may no longer be accessible.}}\hss\hbox{}}\relax
  \let\@evenfoot\@oddfoot}
\makeatother
\newlength{\preprintnoticedrop}
\newlength{\videonoteraise}
\IEEEaftertitletext{\vspace*{-\videonoteraise}{\centering\color{red}\large\bfseries Supplementary video: A video illustrating the XoNav hardware, haptic cues, Wizard-of-Oz trials and autonomous navigation is available at \url{https://doi.org/10.6084/m9.figshare.34701648}\par}\vspace*{6pt}}

\begin{document}

\title{XoNav: An Exoskeleton as a Personalized Haptic Interface for Assisted Navigation}

\author{Debadrata Sarkar, Soumen Sen, Aman Arora \textsuperscript{*}
\thanks{The work was supported by financial grants from the Indian Council of Medical Research (ICMR), New Delhi with file bearing no. EM/Dev/CAR-2024-01-00481/F2/2024, dated: 03/09/2025. This work was also partially supported by the CSIR-National Laboratory Scheme under the ULIP [MLP002613].}
\thanks{This work involved human subjects. Approval of all ethical and experimental procedures was granted by the Institutional Ethical Board of CSIR-CMERI, Durgapur, India (approval number: 1.Adm.(1)/Misc./Committee/97-E, dated: 29/04/2025) and all participants gave written informed consent.}
\thanks{The authors are with the Human Centered Robotics and Cybernetics Group, CSIR-Central Mechanical Engineering Research Institute, Durgapur 713209, India.}
\thanks{ \textsuperscript{*}Corresponding author: Dr Aman Arora (e-mail: aman.cmeri@nic.in)}}

\markboth{}%
{Sarkar \MakeLowercase{\textit{et al.}}: XoNav: An Exoskeleton as a Personalized Haptic Interface for Assisted Navigation}

\maketitle

\begin{abstract}
Walking without vision requires timely, unambiguous guidance, yet many wearable aids encode directions as audio or vibrotactile signals that users must translate into movement. We present XoNav, a four-actuator hip exoskeleton that renders guidance cue directly at the hip joint. It renders Advance cue and two-level Braking cues through sagittal-plane actuators and two-level Steering cues through frontal-plane actuators. A bilateral LSTM-based gait-state classifier uses two thigh-mounted IMUs to identify gait states, enabling haptic cues to be delivered in coordination with the wearer's gait while suppressing them during standing. The classifier generalized to unseen users with recalls of 95.2–97.9\%, with mean detection delays of 38.6 and 63.1 ms for stance-to-swing and swing-to-stance transitions respectively. Constrained human-in-the-loop Bayesian optimization personalized Braking for minimum resistance and Steering for maximum comfort while ensuring detectability and safety. All participants reliably distinguished between grades of their personalized cues and the firmest Braking cue remained below 53\% of the safety limit. In blindfolded Wizard-of-Oz navigation trials along three paths, the participants kept 98.8\% of the walked distance within a 0.75~m wide passage, with a 4.9\% increase in mean progression speed per trial. A body-worn camera, RTAB-Map and Nav2 further enabled a feasibility study of autonomous navigation and detours around previously unmapped obstacles.
\end{abstract}


\section{Introduction}
\IEEEPARstart{N}{avigating} without vision requires guidance about both direction and movement. Canes, guide dogs and electronic aids serve people with visual impairment, while comparable needs arise when sight is obscured or occupied, such as for firefighters in smoke~\cite{slater2025hf} or industrial workers engaged in visually demanding tasks. A wearable RGB-D, audio and vibrotactile system improved navigation for blind and low-vision users~\cite{tang2025nmi}, and bone conduction preserves awareness of environmental sounds~\cite{xue2025nree}. Personalized smartphone audio has also guided blind users outdoors with performance comparable to a human guide~\cite{liu2026nbme}. Robotic canes~\cite{na2024tnsre} and quadruped guides~\cite{chen2023icra} provide physical guidance but occupy a hand and act from outside the body.

Body-worn haptic interfaces offer hands-free guidance, although a survey found that many walking aids still depend on Wizard-of-Oz operators and are not ready for daily use~\cite{kappers2024toh}. Vibrotactile belts reduce collisions during training~\cite{ricci2024jmir}, instrumented walker handles can improve confidence~\cite{lacote2024eh}, and textile interfaces distribute haptic feedback across the body~\cite{dupasquier2024scirobot,park2025ncomms}. Vibrotactile directions nevertheless require interpretation as locomotor actions, while audio competes with ambient cues. Kinesthetic interfaces instead communicate movement through force, as demonstrated by the resistance and pulling rendered by the Kinethreads exosuit~\cite{shen2025uist}.

Lower-limb exoskeletons can convey guidance at the joints responsible for walking. Recent works include multi-activity~\cite{siviy2023nbme} as well as task-agnostic~\cite{molinaro2024nature} joint-moment assistance, data-driven community ambulation~\cite{lee2024sciadv} and online personalization~\cite{kang2025tro}, frontal-plane control of step width and balance~\cite{leestma2024ral,alili2024tbme}, and improved haptic transparency~\cite{kucuktabak2024tro}. Variability in individual responses have motivated human-in-the-loop optimization (HILO)~\cite{slade2024nature,sarkar2026tmrb}, including Bayesian optimization~\cite{ding2018scirobot} and preference-based methods~\cite{lee2023scirobot,ramella2025icra}. Adding vibrotactile signals to exoskeleton torques can modulate walking speed but may increase cognitive load~\cite{wu2024hf}.

Closest to our work, Carrasquillo \emph{et al.}~\cite{carrasquillo2025toh} used artificial potential fields to modulate lower-limb exoskeleton impedance for obstacle avoidance. In virtual reality with occluded vision, this increased clearance and reduced collisions, demonstrating the potential of exoskeleton force feedback for human navigation. However, four gaps remain for real-world wayfinding. First, directional cueing through the hip's frontal-plane degrees of freedom, which govern mediolateral foot placement and heading, remains unexplored. Second, guidance has not been phase-locked to gait events, potentially displacing an unloaded swing limb and overriding, rather than informing, motor intent. Third, navigation cues have used fixed parameters despite individual differences in perception and comfort, while HILO has not been formulated to jointly satisfy detectability, safety and comfort. Finally, reliance on external tracking leaves untested whether egocentric localization from a body-worn camera, despite gait-induced sway, can couple a navigation stack~\cite{macenski2023ras} with a human walker.

\begin{figure*}[!t]
\centering
\includegraphics[scale=0.9]{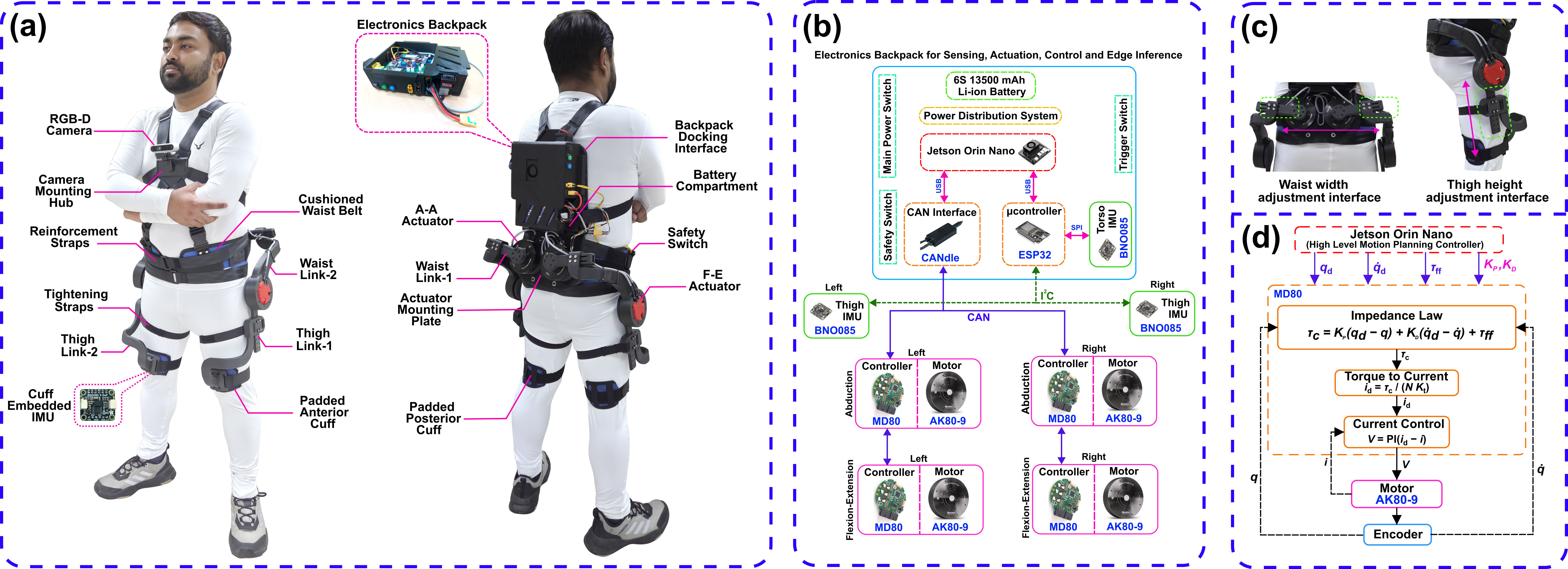}
\caption{XoNav hardware. (a) Front and rear views of the wearable system. (b) Schematic of the electronics architecture showing power distribution, sensing,computation, and actuator communication. (c) Waist-width and thigh-height adjustment mechanisms.
(d) Joint impedance control architecture showing high-level commands from the Jetson and torque and current control executed by the motor driver.}
\label{FIG1}
\end{figure*}

Here, we present XoNav, a hip exoskeleton serving as a personalized haptic interface for assisted navigation (Fig.~\ref{FIG1}). XoNav renders Advance and two-level Braking cues in the sagittal plane and two-level Steering cues in the frontal plane. Each cue is gait-informed, being released at its gait event and withheld while the wearer stands, so that the exoskeleton guides the movements the wearer initiates and never overrides them. The graded cues are personalized for reliable detection at the lowest intensity, within safety limits and in the profile the wearer finds most comfortable. The main contributions are: (i) a four-actuator hip exoskeleton realizing gait-informed haptic navigation cues, timed by a classifier that uses only two thigh IMUs and generalizes to unseen users; (ii) a constrained HILO framework that balances detectability, safety and comfort through a detection constraint, an analytic safety screen and duel-based comfort scoring; (iii) a blindfolded Wizard-of-Oz navigation study, using motion capture and user ratings to evaluate haptic guidance independently of robotic perception and planning errors; (iv) an autonomous navigation feasibility study integrating a body-worn camera with a ROS~2 navigation stack, demonstrating guidance and detours around unmapped obstacles. Our results show that wearers navigate reliably with personalized cues of minimal intensity, a step towards exoskeletons that guide human locomotion rather than driving it.

\section{XoNav Hardware}
XoNav is a bilateral, four-degree-of-freedom hip exoskeleton (Fig.~\ref{FIG1}). Each side has a coaxial flexion--extension (FE) actuator and a posteriorly mounted abduction--adduction (AA) actuator that drives the thigh through waist and thigh links. All four are quasi-direct-drive AK80-9 actuators (CubeMars, China) with 9:1 reduction and MD80 drivers (MAB Robotics, Poland). The low reduction ratio limits backdrive torque, allowing unimpeded walking without cues. Adjustable waist and thigh links and padded cuffs accommodate users and transfer torques (Fig.~\ref{FIG1}(c)). Two BNO085 IMUs (Adafruit Industries, USA) are mounted on the thigh cuffs and a third on the torso. A chest-mounted RealSense D435i (Intel, USA) provides stereo and inertial measurements for navigation. Total system mass is 4.3~kg.

The backpack (Fig.~\ref{FIG1}(b)) contains a 6S 13.5~Ah Li-ion battery, power distribution, an emergency cutoff switch and an event marking trigger switch for data collection. A Jetson Orin Nano (NVIDIA, USA) running ROS~2 Humble performs gait inference, cue generation and navigation. It communicates with the motor drivers over USB-to-CAN (CANdle, MAB Robotics) and receives IMU measurements through an ESP32 (Espressif Systems).

Each joint is driven in impedance mode (Fig.~\ref{FIG1}(d)),
\begin{equation}
\tau_c = K_P\,(q_d - q) + K_D\,(\dot q_d - \dot q) + \tau_{\mathrm{ff}},
\label{eq:impedance}
\end{equation}
Here $q$ and $\dot q$ are the encoder-measured joint angle and velocity and $q_d$ and $\dot q_d$ are their respective references. $K_P$ and $K_D$ denote the virtual stiffness and damping gains, respectively, while $\tau_{\mathrm{ff}}$ is the feed-forward torque. The driver converts $\tau_c$ to a motor-current reference $i_d=\tau_c/(NK_t)$, using gear ratio $N$ and torque constant $K_t$, and tracks it with a PI current loop.At 100~Hz, the Jetson sends $(q_d,\dot q_d,K_P,K_D,\tau_{\mathrm{ff}})$ to each actuator driver, with cue-specific settings of these inputs determining the rendered haptic behavior. A safety node clips position, gain and torque commands to verified limits, including during optimizer evaluations.

\section{Gait-Informed Haptic Cues}
XoNav renders four everyday navigation commands as haptic cues at the hips, namely Advance (keep walking), Braking (slow down or stop), Steering (turn) and Neutral (no guidance) (Fig.~\ref{FIG2}). Their mechanical effects and gait-dependent timing are described below.

\subsection{Cue Vocabulary}
\begin{figure}[!t]
\centering
\includegraphics[scale=0.9]{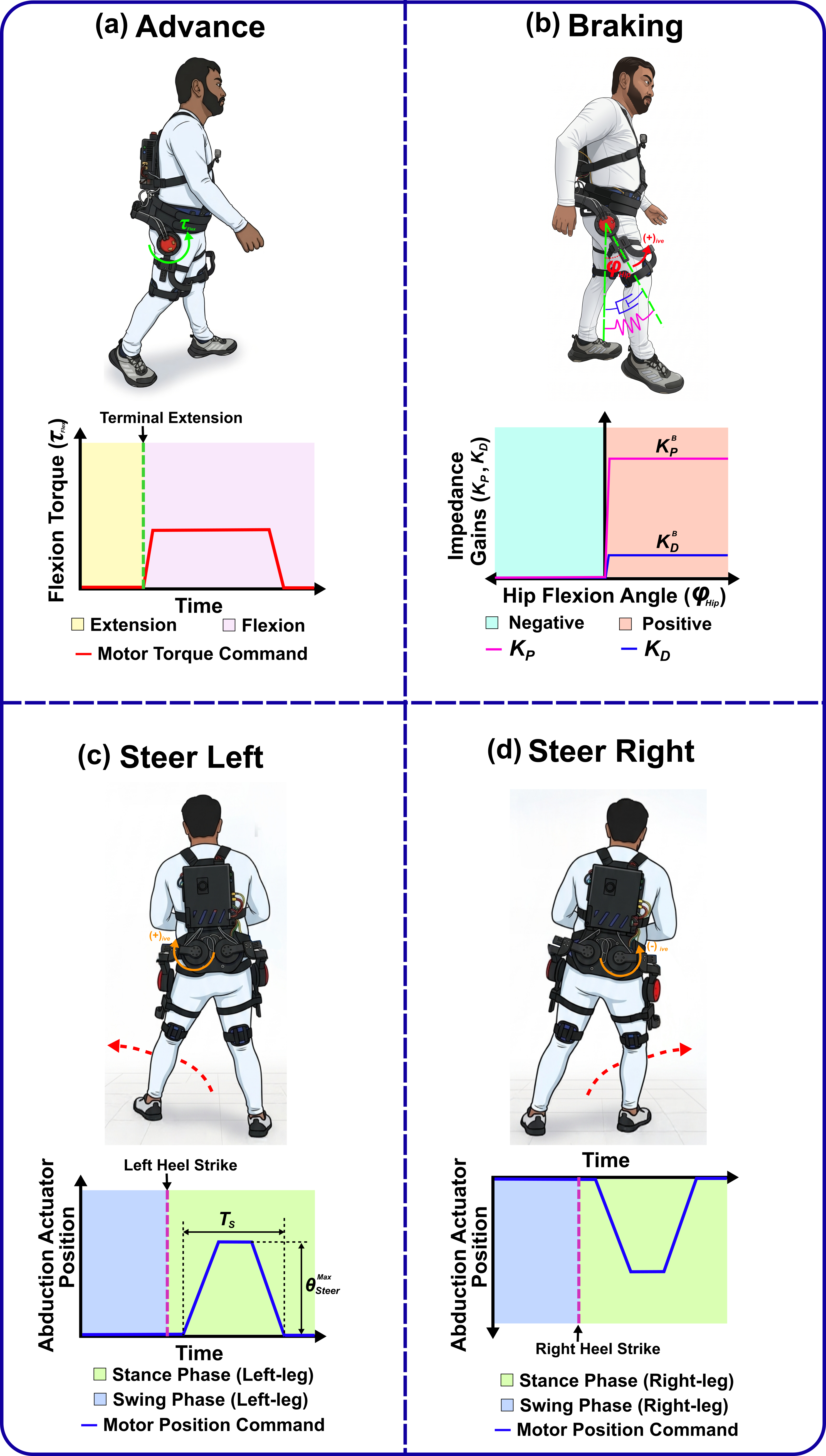}
\caption{Haptic cue profiles. (a) Advance cue.
(b) Braking cue. (c) Left Steering cue.
(d) Right Steering cue.}
\label{FIG2}
\end{figure}

The \emph{Advance} cue encourages continued walking. At terminal hip extension, identified by the extension-to-flexion zero crossing of thigh angular velocity, the FE actuator produces a brief trapezoidal flexion torque $\tau_{\mathrm{ff}}$ with $K_P=K_D=0$ (Fig.~\ref{FIG2}(a)). The perceived gentle forward pull accompanies the wearer's ensuing step rather than initiating one.

The \emph{Braking} cue asks the wearer to slow down or to stop. The FE actuator renders a virtual spring--damper about the neutral hip angle ($q_d = \dot q_d = 0$) whose gains are switched by the hip flexion angle $\varphi$ measured by the actuator encoder,
\begin{equation}
(K_P, K_D) =
\begin{cases}
(\Kb,\; K_D^{\mathrm{B}}), & \varphi > 0,\\
(0,\; 0), & \varphi \le 0.
\end{cases}
\label{eq:braking}
\end{equation}
The wearer feels resistance during forward thigh swing, but extension, standing and backward stepping remain unimpeded (Fig.~\ref{FIG2}(b)). Slow-down uses $\Kb=K_0$, Stop uses $\Kb=K_0+\Delta K$, and damping $K_D^{\mathrm{B}}$ remains fixed.

The \emph{Steering} cue abducts the turning-side thigh through its AA actuator with a trapezoidal profile of duration $T_s$ and peak angle $\theta_{\mathrm{Steer}}^{\max}$. Rise, hold and return occupy $0.3T_s$, $0.4T_s$ and $0.3T_s$, respectively (Fig.~\ref{FIG2}(c), (d)). Mirrored left and right actuation produces an outward nudge toward the requested turn. Gentle Steering uses $\theta_{\mathrm{Steer}}^{\max}=\theta_0=15^\circ$ for mild directional corrections, while Sharp Steering uses $\theta_{\mathrm{Steer}}^{\max}=\theta_0+\Delta\theta$ for more pronounced turns.

In \emph{Neutral}, FE gains are zero for free backdriving. The AA joints hold a wearer-selected neutral angle set at donning with nominal impedance gains, sufficient to support the FE actuators and links but compliant enough to follow frontal-plane motion during walking.

\subsection{Gait-Informed Rendering}
XoNav times its cues to accompany wearer-initiated movement rather than initiating it. Steering begins immediately during turning-side stance or waits until the next heel strike if commanded during swing, after which the full profile executes. With the foot loaded, the abduction cue is conveyed through the pelvis and trunk rather than displacing the swing foot and pilot wearers found it more perceptible than during swing. All cues are suppressed during standing.

A bilateral classifier estimates Standing, Stance and Swing for each leg using only two thigh-IMU pitch signals sampled at 100~Hz. Every 50~ms, seven causal features (mean, RMS, maximum, minimum, range, net change and standard deviation) are computed for each signal over the preceding 300~ms. Six consecutive feature vectors (550~ms total history) feed a shared long short-term memory (LSTM) layer with separate classification heads, requiring two consecutive identical predictions to confirm a state transition. Heel and toe force-sensitive resistors provide stance and swing labels, while event triggers mark standing intervals, both used only for training.

\begin{figure}[!b]
\centering
\includegraphics[scale=0.7]{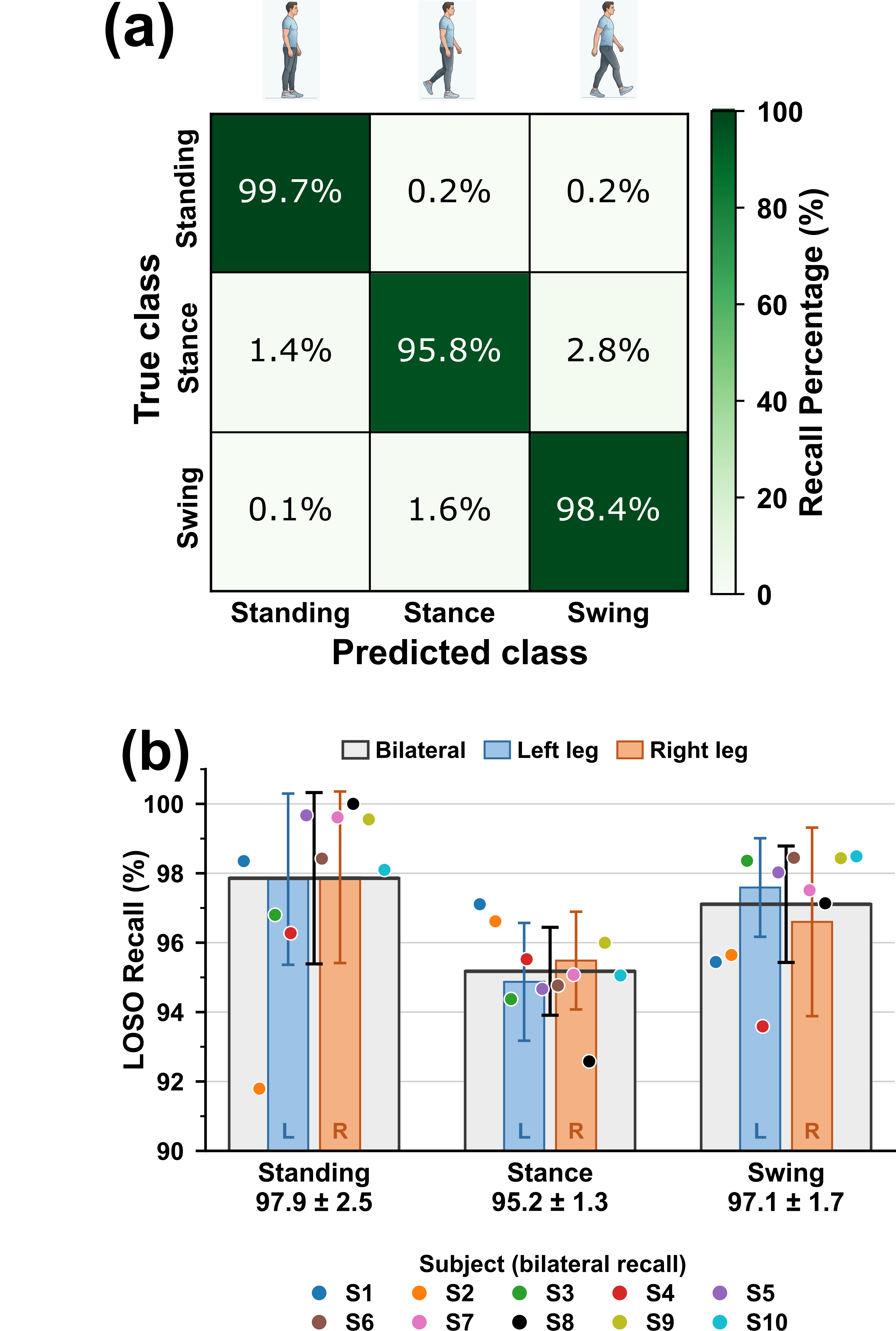}
\caption{Bilateral gait-state classifier. (a) Validation confusion matrix of the deployed model (recall, \%). (b) Leave-one-subject-out recall for each class, bilateral and per leg; dots show individual participants, and the labels give the bilateral mean $\pm$ SD.}
\label{FIG3}
\end{figure}

The classifier was trained on recordings from ten participants.
The deployed model achieved recalls of 99.7\%, 95.8\% and
98.4\% for Standing, Stance and Swing, respectively
(Fig.~\ref{FIG3}(a)). Leave-one-subject-out (LOSO)
validation, avoiding leakage from overlapping windows and
assessing unseen-user generalization, yielded bilateral recalls
of $97.9\pm2.5\%$, $95.2\pm1.3\%$ and $97.1\pm1.7\%$,
respectively, with comparable performance across legs
(Fig.~\ref{FIG3}(b)). On the Jetson, the classifier
ran at 20~Hz with an inference latency of $18.2\pm4.1$~ms.
Before two-prediction confirmation, transition detection delays
were $38.6\pm35.3$~ms (stance-to-swing) and
$63.1\pm31.8$~ms (swing-to-stance). Standing-to-swing
transitions took $37.5\pm25.0$~ms, while transitions to
Standing took $41.4\pm117.3$~ms from Stance and
$90.0\pm108.4$~ms from Swing. These delays permit prompt gait-dependent cue rendering and suppression.

\section{Human-in-the-Loop Cue Personalization}
\label{sec:hil}
\begin{figure*}[!t]
\centering
\includegraphics[scale=0.9]{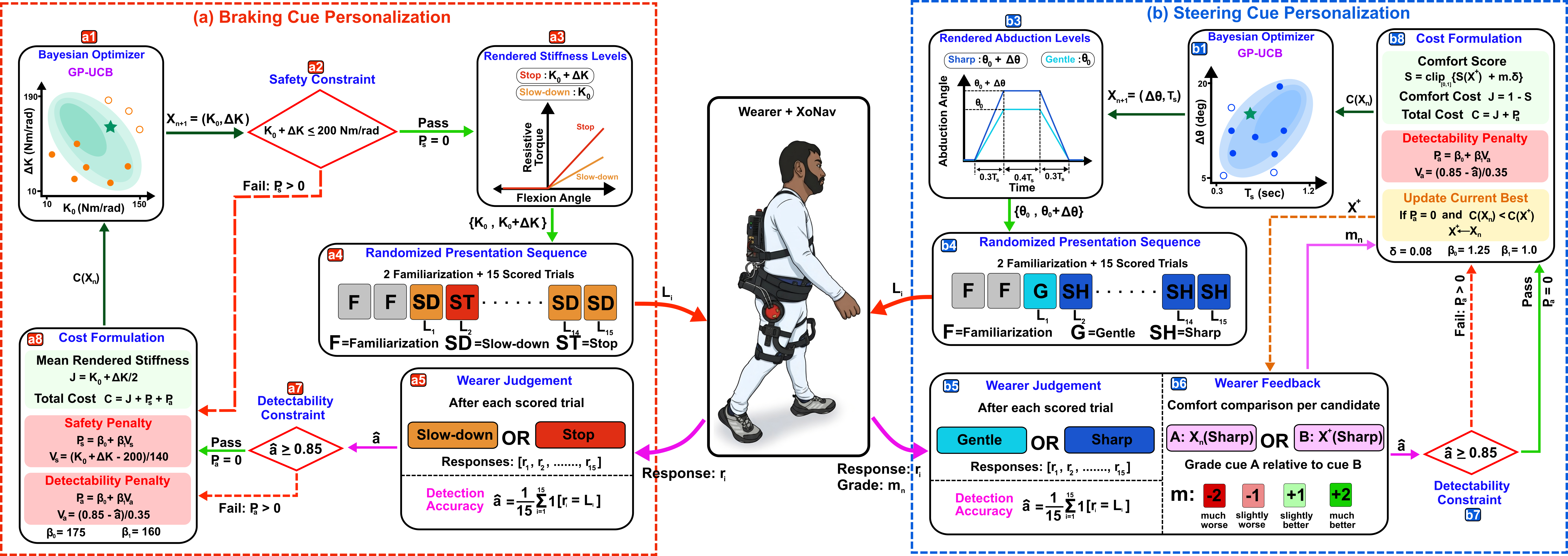}
\caption{Human-in-the-loop optimization of (a) Braking and (b) Steering. The optimizer proposes a candidate (a1, b1); Braking undergoes a safety check (a2). Randomized cue presentations (a3--a4, b3--b4) are identified by the wearer (a5, b5). For Steering, the new Sharp cue is also compared with the incumbent (b6). Detection constraints (a7, b7) determine the penalized cost (a8, b8), returned to the optimizer. Dashed red paths indicate penalties.}
\label{FIG4}
\end{figure*}

Haptic cues must be reliably distinguishable without
unnecessarily interfering with natural walking, yet wearers
differ in their perception of hip forces and motions and
in comfort preferences. XoNav therefore personalizes its
graded cues through constrained human-in-the-loop Bayesian
optimization, using wearer feedback to balance detectability,
comfort and safety (Fig.~\ref{FIG4}). Both loops share
the evaluation and constraint-handling procedures described
below. Sessions lasted 30--50~min, with rest breaks on request.

\subsection{Personalization Framework}

Each candidate parameter set $X$ is evaluated in a walking
block comprising two familiarization presentations followed
by 15 randomized scored presentations (Fig.~\ref{FIG4},
a3--a4, b3--b4),
during which the participant identifies the cue level after
each presentation (a5, b5). The resulting detection accuracy
$\hat a$ must satisfy $\hat a\ge0.85$ (at least 13/15 correct;
chance probability below 0.005) (a7, b7). Constraint violations
are penalized in the candidate cost (a8, b8) as
\begin{equation}
\begin{aligned}
C &= J + \textstyle\sum_j P_j,\\
P_j &= \begin{cases}
\beta_0 + \beta_1 \min\{1, V_j\},
& \text{if constraint } j \text{ is violated},\\
0, & \text{otherwise},
\end{cases}
\end{aligned}
\label{eq:cost}
\end{equation}
where $V_j$ is the normalized constraint violation,
$\beta_0$ separates feasible from infeasible costs and
$\beta_1$ scales the penalty with violation magnitude.
A Gaussian process with a Mat\'ern-5/2 kernel models
the cost and the next candidate minimizes the lower
confidence bound (a1, b1)
\begin{equation}
X_{n+1} = \arg\min_X \;
\mu_n(X)-\kappa\,\sigma_n(X), \qquad \kappa=4,
\label{eq:ucb}
\end{equation}
where $\mu_n$ and $\sigma_n$ are the posterior mean and
standard deviation of $C$, respectively. Each session
starts with five random candidates and the best evaluated
feasible candidate is selected.

\subsection{Braking Cue Personalization}

Braking personalization seeks the lowest resistance that
allows wearers to reliably distinguish the Slow-down and
Stop cues, thereby minimizing interference with natural
walking. The parameter set $X=(K_0,\Delta K)$, with
$K_0\in[10,150]$ and $\Delta K\in[10,190]$~N$\cdot$m/rad,
minimizes the mean commanded stiffness
$J=K_0+\Delta K/2$. Alongside the detection constraint
$\hat a\ge0.85$, the Stop stiffness is bounded by
$K_0+\Delta K\le200$~N$\cdot$m/rad. Candidates violating
this known safety bound are screened and penalized before
rendering (a2), preventing their
application to the wearer. The penalties use $\beta_0=175$
and $\beta_1=160$, with 20 evaluations per session
(five random, 15 guided). Two Braking levels were adopted
after pilot trials showed that an intermediate third level
could not be reliably distinguished from its neighbors.

\subsection{Steering Cue Personalization}

Steering personalization seeks the most comfortable Sharp
cue that remains reliably distinguishable from Gentle
Steering. With Gentle amplitude fixed at $\theta_0=15^\circ$,
the Sharp cue is optimized over $X=(\Delta\theta,T_s)$,
where $\Delta\theta\in[5^\circ,30^\circ]$ and
$T_s\in[0.3,1.2]$~s. No additional safety constraint is
imposed, as all amplitudes remain within the AA joint range.
Since pilot participants found absolute comfort ratings
inconsistent, each candidate is evaluated through a
relative comparison with the current best $X^+$.
After the identification block, the incumbent Sharp cue
is replayed before the candidate's, which is rated as
much worse, slightly worse, slightly better or much better,
encoded as $m\in\{-2,-1,+1,+2\}$
(b6).
This comparison updates a chained comfort score:
\begin{equation}
S(X_n) = \operatorname{clip}_{[0,1]}\!
\left(S(X^+) + m\,\delta\right),
\label{eq:duel}
\end{equation}
with $\delta=0.08$ and $S=0.5$ for the first admissible
candidate. The corresponding optimization cost is:
\begin{equation}
J = 1-S(X_n).
\label{eq:steering_cost}
\end{equation}

A candidate replaces $X^+$ only if admissible
and bears a lower-cost. Detection penalties use $\beta_0=1.25$
and $\beta_1=1.0$, with 15 evaluations per session
(five random, ten guided).

\begin{figure}[!t]
\centering
\includegraphics[scale=0.85]{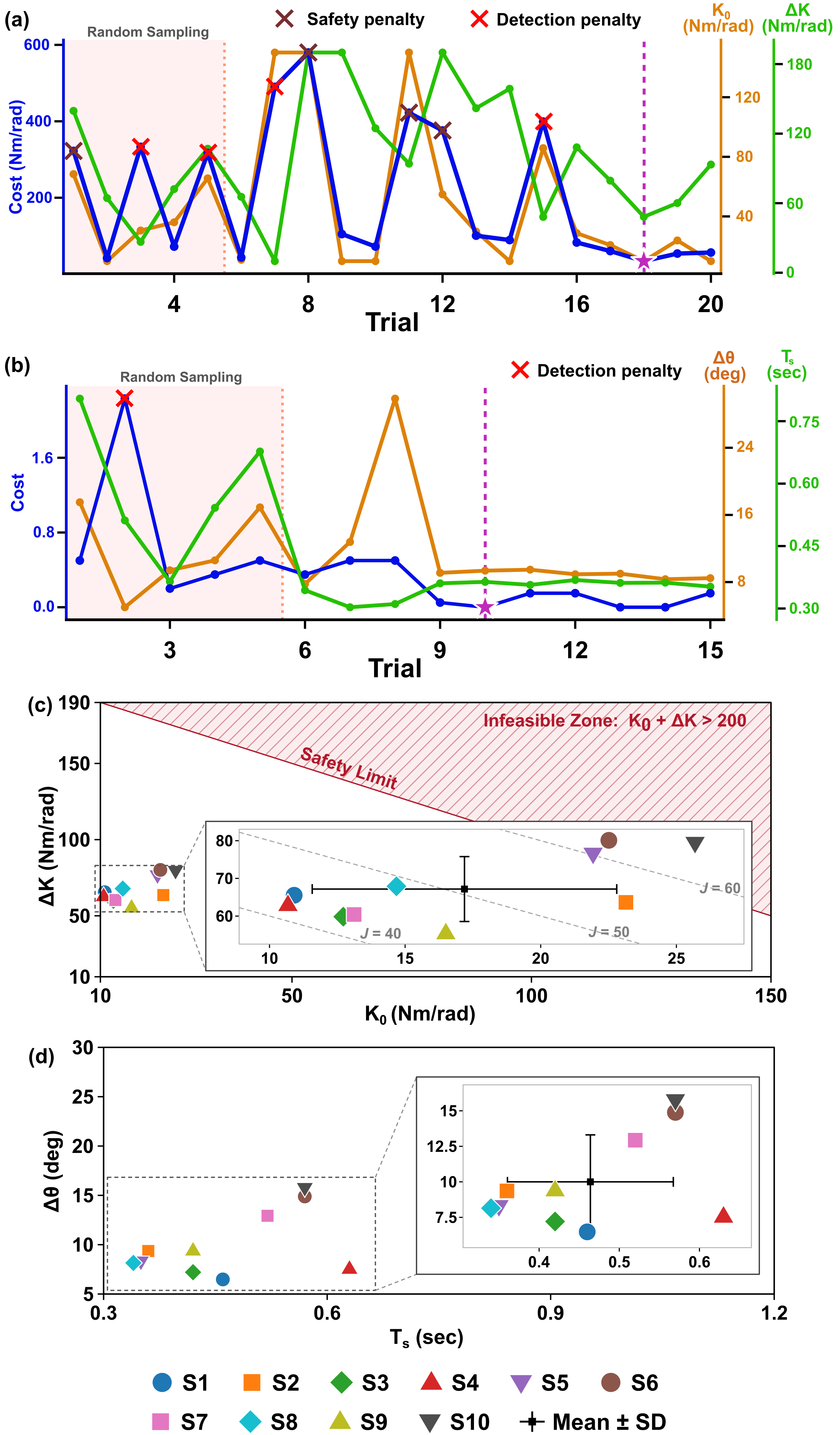}

\caption{HIL personalization. (a), (b) Representative Braking
and Steering optimization histories showing costs and
parameters; crosses mark penalized candidates and stars
the selected settings. (c), (d) Personalized settings
across ten participants with magnified insets (mean $\pm$ SD).
Dashed lines in (c) indicate iso-cost contours of
$J=K_0+\Delta K/2$.}

\label{FIG5}
\end{figure}

\subsection{Results}
In representative sessions (Fig.~\ref{FIG5}(a), (b)), infeasible Braking candidates were penalized either before rendering for exceeding the safety limit or afterward for poor detectabiility. The optimizer then moved toward low base stiffness, selecting the best candidate at evaluation 18. 
For Steering, a small amplitude increment failed the detection criterion and was penalized, while the best candidate was selected at evaluation 10. Clipping the chained comfort score to $[0,1]$ prevents unbounded accumulation across comparisons and keeps its cost below the penalty offset. Although the score thereby saturates, ties at the ceiling retain the current best, deliberately favoring a stable selection over differences that pilot wearers could not reliably perceive between similarly comfortable profiles.

Across ten participants (Fig.~\ref{FIG5}(c), (d)), Braking converged to low Slow-down stiffness ($K_0=10.7$--25.7~N$\cdot$m/rad) and substantially higher Stop stiffness ($\Delta K=55.5$--80.1~N$\cdot$m/rad), a Stop-to-Slow-down ratio of $5.2\pm1.1$. Relative contrast mattered more than absolute stiffness. In one representative session, a base near 85~N$\cdot$m/rad failed detection even though a similar increment passed at a base near 10~N$\cdot$m/rad. This agrees with Weber's law: discriminable increments rise with baseline stimulus intensity. The resulting Stop stiffness was 72--105~N$\cdot$m/rad, no more than 53\% of the safety limit. The consistent optima suggest $K_0\approx17$ and $\Delta K\approx67$~N$\cdot$m/rad as starting values for shorter future sessions.

Steering preferences varied more widely, with $\Delta\theta=6.5$--15.7$^\circ$ (Sharp amplitudes 1.4--2.0 times Gentle) and $T_s=0.34$--0.63~s. Preferred profiles were brief enough for single-stance rendering at the tested walking speeds. Participants favoring larger amplitudes also favored longer durations, suggesting sensitivity to abduction rate. Frontal-plane displacement was discriminated at smaller relative contrasts than sagittal-plane resistance. Thus, Braking showed a shared low-intensity pattern, whereas Steering required individual comfort tuning; all selected settings met the detection criterion $\hat a\ge0.85$.

\section{Wizard-of-Oz Navigation Study}

To evaluate navigation using the personalized cues independently of perception, localization and planning errors, we adopted a Wizard-of-Oz protocol~\cite{kappers2024toh}. A trained operator issued navigation commands, while the untethered exoskeleton rendered the corresponding gait-informed cues as in autonomous
operation.

\subsection{Participants and Protocol}

\begin{figure*}[!t]
\centering
\includegraphics[scale=0.8]{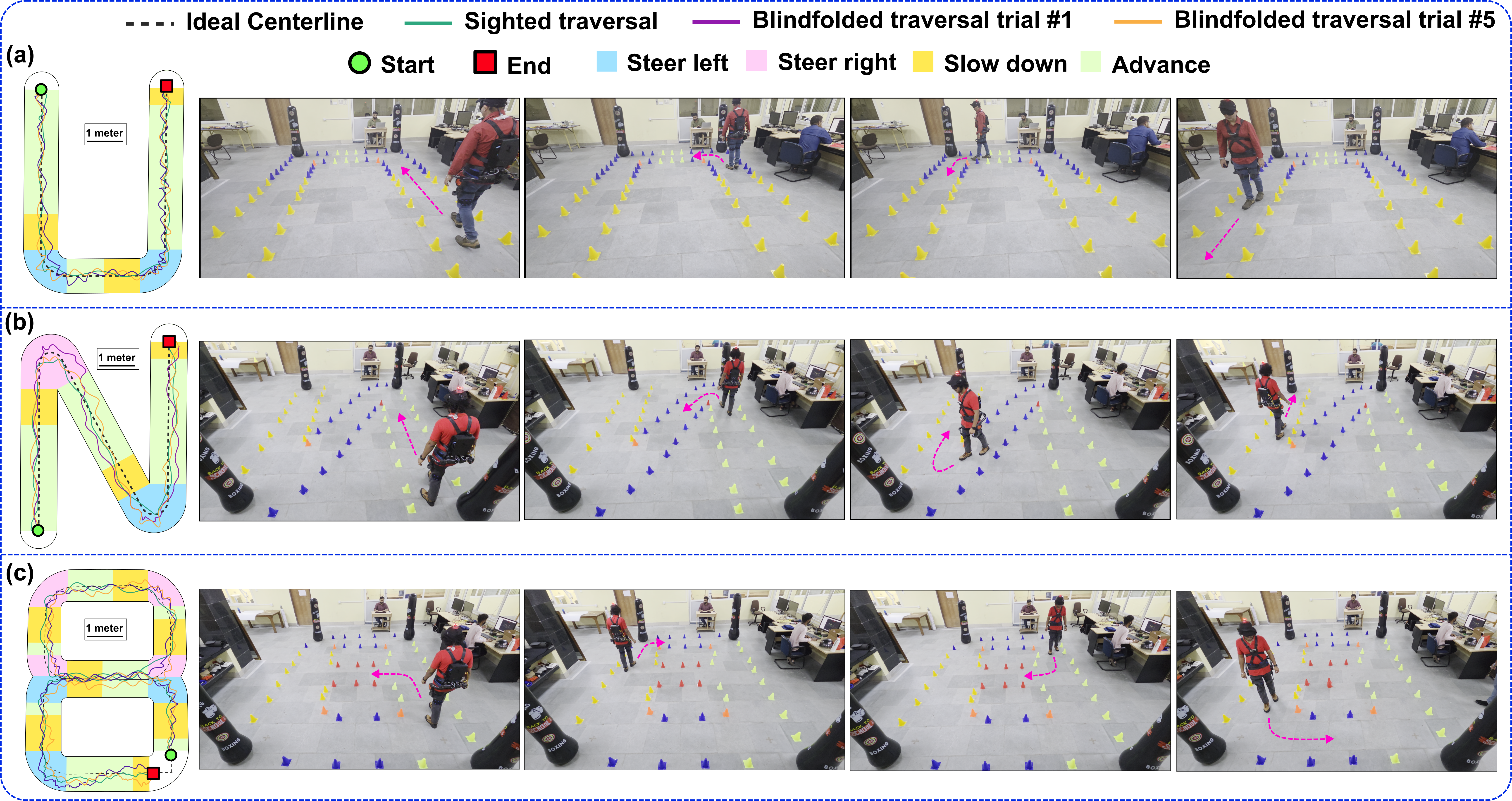}

\caption{Wizard-of-Oz navigation trials on (a) U-,
(b) N- and (c) 8-shaped paths. Left: centerline,
cue zones and traversed paths of one participant
during sighted baseline and blindfolded trials 1 and 5.
Right: snapshots captured at different instants
during a blindfolded trial.}

\label{FIG6}
\end{figure*}

Ten healthy male participants (age $28.1\pm4.9$~years;
height $166.3\pm5.9$~cm; mass $67.3\pm8.7$~kg;
eight right-leg dominant) navigated blindfolded using
their personalized cues. U-, N- and 8-shaped paths of
10.5, 12.5 and 17~m were marked as 0.75~m-wide passages,
approximately 1.5 times the combined hip width of the
wearer and exoskeleton (Fig.~\ref{FIG6}).
An operator in the arena issued Advance, Braking and
Steering commands at predefined cue zones using a
Bluetooth keyboard connected to the onboard computer,
avoiding network-based ROS~2 teleoperation delays.
Each participant completed one sighted baseline per path
to characterize natural head sway, followed by five
blindfolded trials per path (150 guided trials).
A motion-capture system (PhaseSpace, USA) tracked
a head-mounted LED marker throughout.

\subsection{Metrics and Analysis}
Trajectories were referenced to each path's centerline. Progression speed is centerline advancement divided by time; corridor adherence is the fraction of walked distance inside the passage. Lateral wander accumulates head displacement side to side per metre of progress, capturing weaving even when mean offset is small. Corrective reversals count per-metre lateral-drift changes exceeding 50~mm, and RMS heading error measures deviation from path heading. To remove baseline head sway and marker variability, each participant's same-path sighted value was subtracted from wander and, in quadrature, from heading error.

Adaptation across trials was assessed using a linear
mixed-effects model of log-transformed metrics, with
trial number and path as fixed effects and a
participant-by-path random intercept. The trial
coefficient estimated percentage change per repetition,
accounting for individual and path differences.

\subsection{Results}

\begin{figure*}[!t]
\centering
\includegraphics[scale=0.9]{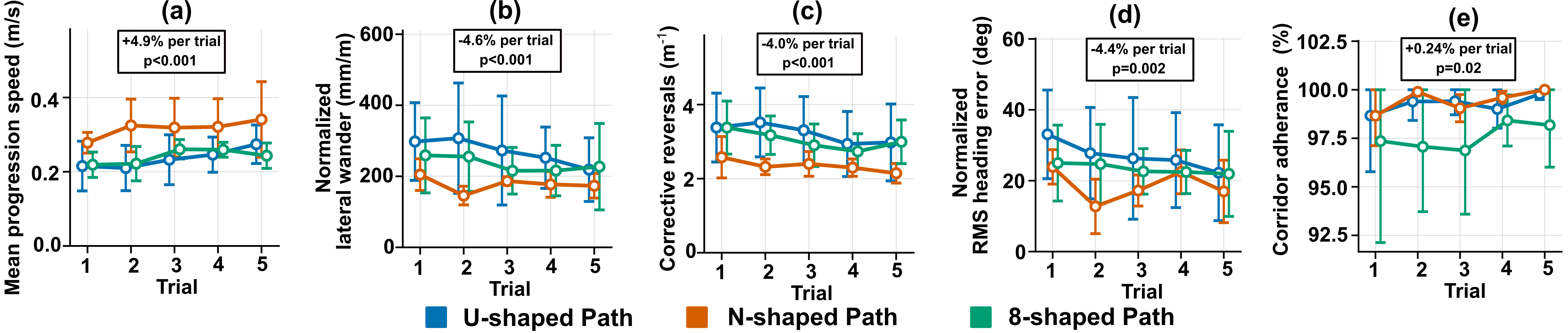}
\caption{Changes over five blindfolded trials (mean $\pm$ SD per path): (a) progression speed, (b) normalized lateral wander, (c) corrective reversals, (d) normalized RMS heading error and (e) corridor adherence. Insets report mixed-effects model percentage change per trial and $p$-values.}
\label{FIG7}
\end{figure*}

All 150 trials were completed, with mean corridor adherence
of 98.8\% and 93\% of trials exceeding 95\% adherence.
Adherence improved by 0.24\% per trial ($p=0.02$;
Fig.~\ref{FIG7}(e)), while mean progression speed
rose from 0.24 to 0.29~m/s, a modeled increase of 4.9\%
per trial ($p<0.001$; Fig.~\ref{FIG7}(a)).
Concurrently, normalized lateral wander, heading error
and corrective reversals decreased by 4.6\% ($p<0.001$),
4.4\% ($p=0.002$) and 4.0\% ($p<0.001$) per trial,
respectively (Fig.~\ref{FIG7}(b)--(d)), while
RMS lateral deviation remained near 154~mm.
Adherence was 99.3\% on U, 99.4\% on N and 97.6\%
on the turn-intensive 8-shaped path. Overall, repeated
haptic guidance improved speed and directional stability
while maintaining high corridor adherence.

\begin{figure*}[!t]
\centering
\includegraphics[scale=0.9]{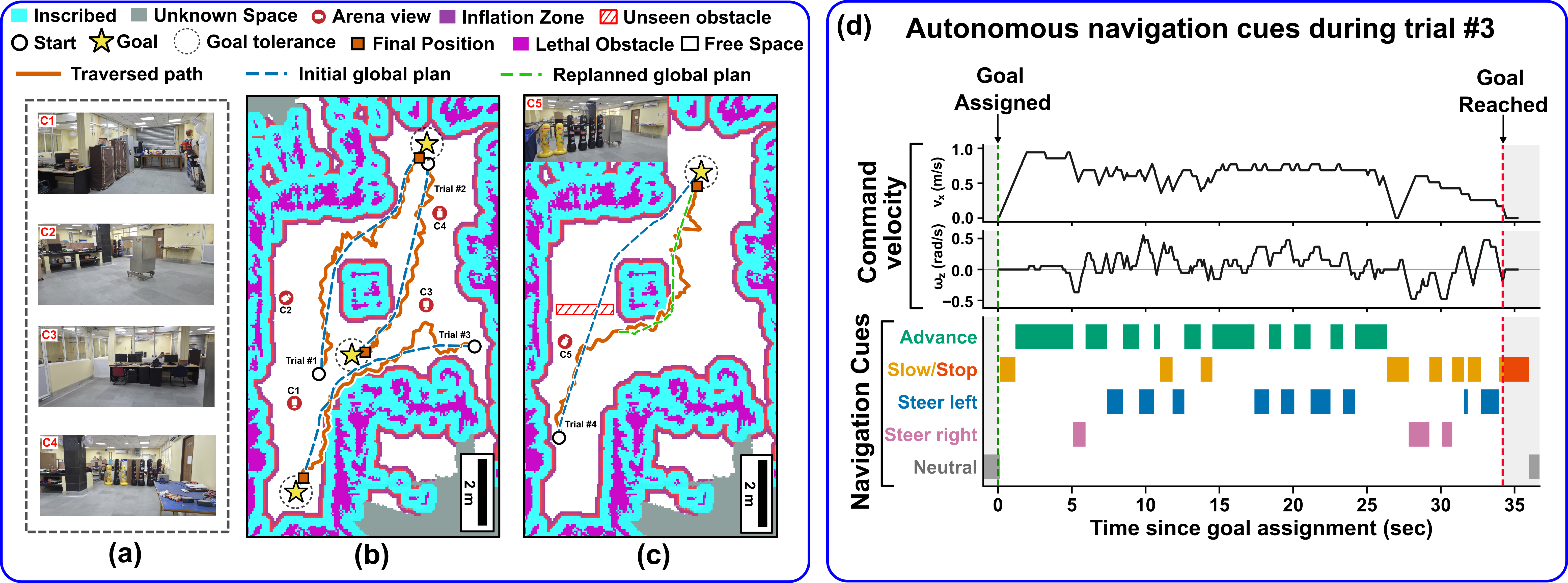}
\caption{Autonomous navigation. (a) Arena views (C1--C4). (b) Costmap, traversed paths and initial global plans for three nominal trials. (c) Replanned route around an unmapped obstacle (C5). (d) Velocity commands and rendered cues from goal assignment to arrival in nominal trial 3.}
\label{FIG8}
\end{figure*}

\section{Autonomous Navigation Feasibility}
A chest-mounted camera sways and changes tilt and height with the wearer's gait, complicating localization compared with wheeled platforms. Since prior exoskeleton navigation was evaluated with external tracking in virtual environments~\cite{carrasquillo2025toh}, we tested whether onboard sensing alone could support closed-loop XoNav guidance.

\subsection{Localization and Navigation Stack}

The furnished laboratory arena was mapped using RTAB-Map
with the chest-mounted D435i in stereo-inertial mode.
Infrared stereo images ($848\times480$~px, 30~Hz;
emitter off) provided visual odometry, while the camera
IMU, fused through a Madgwick filter, provided
gravity-aligned orientation. During navigation, RTAB-Map
localized against the saved map, using the camera frame
as the wearer reference so that heading and look-ahead
were evaluated relative to the trunk.

Nav2~\cite{macenski2023ras} used 0.05~m-resolution
costmaps combining the static map, live depth-point-cloud
obstacles (marking 0.3--3.0~m, clearing to 4.0~m)
and inflation (0.45~m radius, scaling factor 5.0)
around a 0.30~m-radius circular footprint.
A Dijkstra-based NavFn global planner generated paths,
while a Dynamic Window Approach-based (DWB) controller
tracked them at 10~Hz within an $8\times8$~m rolling
window. Its configuration included human-compatible
acceleration limits of 0.5~m/s$^2$ and
1.0~rad/s$^2$, a 3~s trajectory simulation horizon
and a 1.2~m/s speed ceiling.

To avoid unsafe maneuvers during vision-occluded
walking, recovery was limited to costmap clearing
and waiting rather than rotation or reversing.
Goal arrival required only a position tolerance
of 0.5~m, since trunk heading oscillates during
walking.

\subsection{From Velocity Commands to Haptic Cues}

A ROS~2 node converts smoothed Nav2 velocity commands
$(v_x,\omega_z)$ into discrete haptic cues. Steering takes
priority when $|\omega_z|>0.25$~rad/s and persists until
it falls below 0.15~rad/s, with direction set by the
sign of $\omega_z$. Otherwise, $v_x\ge0.55$~m/s triggers
Advance, retained until $v_x<0.45$~m/s, while lower
forward velocities trigger Slow-down. Hysteresis and
a minimum cue duration of 0.8~s prevent chattering,
with Braking and Neutral exempted for prompt response.
Unlike velocity-derived Slow-down, Stop overrides other
cues upon goal arrival or obstacle entry into the Nav2
collision monitor's stop zone. It remains active until
the classifier detects standing, after which Neutral
is activated. When Stop is inactive, Neutral is also
activated if no velocity commands are received for
0.75~s. The velocities convey direction and urgency
rather than impose walking speeds.

\subsection{Trials and Outcomes}

Three participants each completed five nominal start-to-goal
trials and three trials with an obstacle placed along the
planned route but absent from the saved map, totaling
24 trials. Fig.~\ref{FIG8}(b) shows representative
nominal paths of one participant, while
Fig.~\ref{FIG8}(c) illustrates an obstacle trial in
which the live point cloud registered the added obstacle
and Nav2 replanned around the central island, allowing
the participant to follow the cues safely to the goal.
The cue sequence of a nominal trial
(Fig.~\ref{FIG8}(d)) shows Advance predominating
along straight segments, Steering at turns and Slow-down
during the final approach.

Fourteen of 15 nominal trials reached the goal, although
two involved light furniture contact followed by recovery
without assistance. The remaining trial failed when sudden
localization drift disrupted planning. Similarly, eight of
nine obstacle trials successfully followed replanned detours,
while delayed obstacle registration in the remaining trial
prevented replanning and resulted in a collision. Across
all trials, walking speed averaged $0.26\pm0.015$~m/s,
with a planned-to-walked path-length ratio of
$69.6\pm5.0\%$. Lateral sway and corrective movements
may have contributed to this difference. Overall, these
results demonstrate the feasibility of autonomous haptic
guidance using body-worn sensing under the tested indoor
conditions, while identifying localization drift and
delayed obstacle detection as important limitations.


\section{User Feedback}

\begin{figure}[!t]
\centering
\includegraphics[scale=1.0]{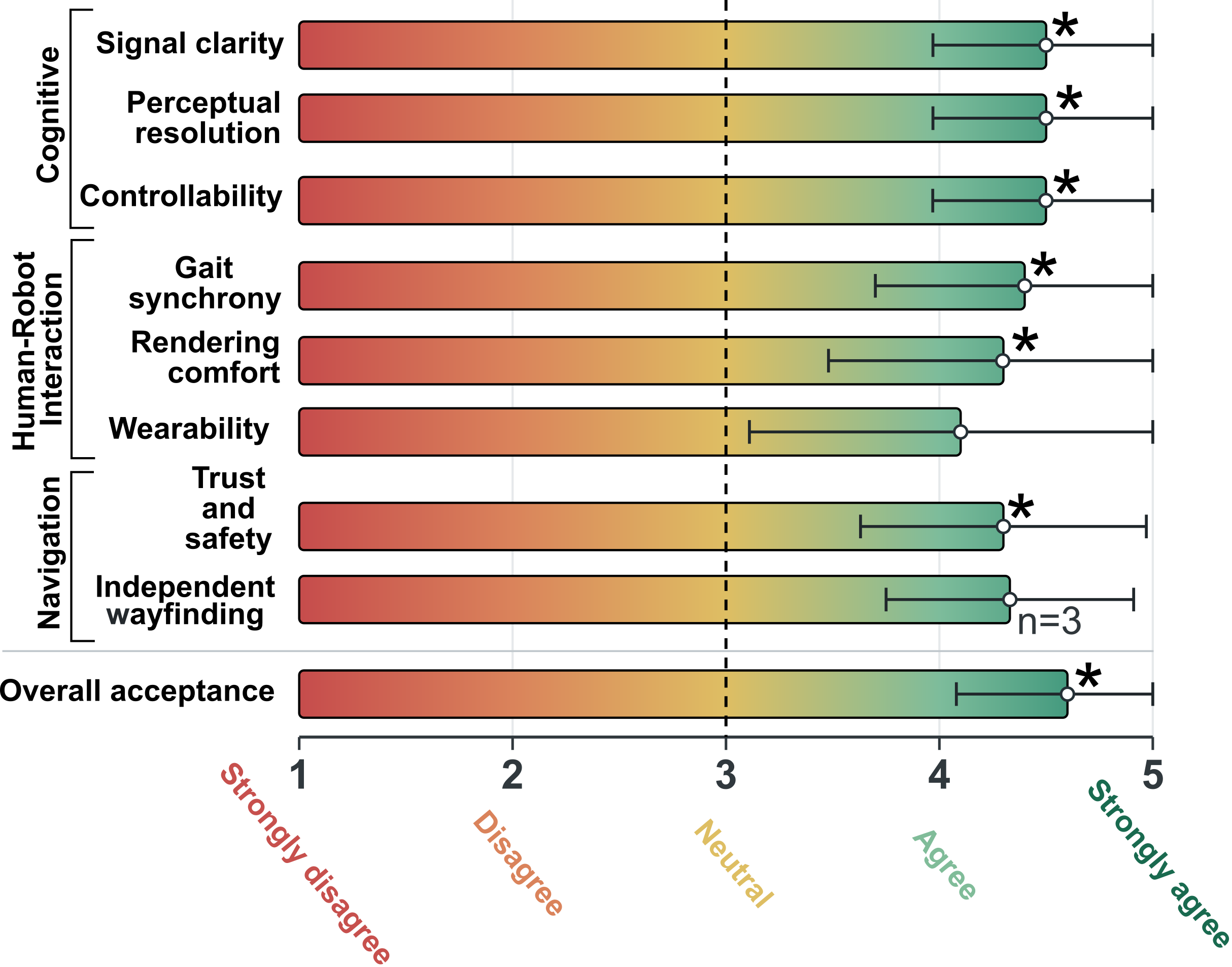}

\caption{Post-session questionnaire ratings
($n=10$; independent wayfinding $n=3$).
Markers and error bars indicate mean $\pm$ SD on a
five-point agreement scale. Asterisks denote
significance above neutral (two-sided exact sign
tests, Holm-adjusted $p<0.05$).}

\label{FIG9}
\end{figure}

Participants rated nine aspects of XoNav after the
experiments (Fig.~\ref{FIG9}).
High ratings for cue clarity, distinguishability and
control over walking supported the gait-informed
guidance approach, while cue timing and comfort
also scored above neutral. Wearability showed the
greatest variation, suggesting scope for reducing
device mass and improving fit. Trust and perceived
safety remained favorable despite blindfolding.
The three autonomous-navigation participants rated
independent wayfinding positively, with overall
acceptance receiving the highest rating.

\section{Conclusion}

XoNav demonstrates how a hip exoskeleton can serve as a
personalized, gait-informed haptic interface that conveys navigation
guidance directly through joint motion rather than symbolic
audio or vibration. Its four actuators render Advance,
two-level Braking and Steering cues across the sagittal and
frontal hip planes, coordinated with wearer-initiated movement
through a classifier using only two thigh IMUs. The classifier
achieved 95.2--97.9\% leave-one-subject-out recall, with gait
transitions detected within tens of milliseconds, enabling
gait-dependent cue delivery and suppression during standing.

Constrained human-in-the-loop optimization yielded reliably
distinguishable Braking cues with minimal resistance and
individually comfortable Steering cues across ten participants,
with maximum Stop stiffness remaining below 53\% of the
safety bound. This personalization allowed guidance without
unnecessarily restraining natural walking. In 150 blindfolded
Wizard-of-Oz trials, participants maintained 98.8\% corridor
adherence within a 0.75~m passage while improving progression
speed by an estimated 4.9\% per trial. The same cues also
supported autonomous navigation using a body-worn camera,
RTAB-Map and Nav2, including detours around unmapped
obstacles. Together, these findings demonstrate the potential
of personalized kinesthetic guidance for assisted navigation,
although validation remains limited to blindfolded sighted
participants in controlled indoor environments. Future work
will evaluate XoNav with people with visual impairment
against white-cane navigation, improve device wearability
and extend autonomy to less structured environments.

A supplementary video illustrates the XoNav hardware,
control architecture and haptic cues, alongside blindfolded
Wizard-of-Oz trials and camera-based autonomous indoor
navigation.


\begin{thebibliography}{99}
\bibitem{slater2025hf}
A.~E. Slater \emph{et al.}, ``Navigating in zero-visibility: A haptic guidance system for improving egress and situation awareness of professional firefighters,'' \emph{Hum. Factors}, vol.~67, no.~11, pp.~1152--1169, 2025, doi: 10.1177/00187208251348020.

\bibitem{tang2025nmi}
J.~Tang \emph{et al.}, ``Human-centred design and fabrication of a wearable multimodal visual assistance system,'' \emph{Nat. Mach. Intell.}, vol.~7, no.~4, pp.~627--638, 2025, doi: 10.1038/s42256-025-01018-6.

\bibitem{xue2025nree}
W.~Xue \emph{et al.}, ``Interactive wearable digital devices for blind and partially sighted people,'' \emph{Nat. Rev. Electr. Eng.}, vol.~2, no.~6, pp.~425--439, 2025, doi: 10.1038/s44287-025-00170-w.

\bibitem{liu2026nbme}
R.~Liu and P.~Slade, ``Improving outdoor navigation for people with blindness using an AI-driven smartphone application and personalized audio guidance,'' \emph{Nat. Biomed. Eng.}, early access, Aug. 2026, doi: 10.1038/s41551-026-01772-x.

\bibitem{na2024tnsre}
Q.~Na, H.~Zhou, H.~Yuan, M.~Gui, and H.~Teng, ``Improving walking path generation through biped constraint in indoor navigation system for visually impaired individuals,'' \emph{IEEE Trans. Neural Syst. Rehabil. Eng.}, vol.~32, pp.~1221--1232, 2024, doi: 10.1109/TNSRE.2024.3375944.

\bibitem{chen2023icra}
Y.~Chen \emph{et al.}, ``Quadruped guidance robot for the visually impaired: A comfort-based approach,'' in \emph{Proc. IEEE Int. Conf. Robot. Autom. (ICRA)}, 2023, pp.~12078--12084, doi: 10.1109/ICRA48891.2023.10160854.

\bibitem{kappers2024toh}
A.~M.~L. Kappers, R.~J. Holt, T.~J.~W. Junggeburth, M.~Fa~Si~Oen, B.~J.~T. van~de~Wetering, and M.~A. Plaisier, ``Hands-free haptic navigation devices for actual walking,'' \emph{IEEE Trans. Haptics}, vol.~17, no.~4, pp.~528--545, 2024, doi: 10.1109/TOH.2024.3405551.

\bibitem{ricci2024jmir}
F.~Ricci, L.~Liguori, E.~Palermo, J.~Rizzo, and M.~Porfiri, ``Navigation training for persons with visual disability through multisensory assistive technology: Mixed methods experimental study,'' \emph{JMIR Rehabil. Assist. Technol.}, vol.~11, Art. no. e55776, 2024, doi: 10.2196/55776.

\bibitem{lacote2024eh}
I.~Lac\^{o}te, P.-A. Cabaret, C.~Pacchierotti, M.~Babel, D.~Gueorguiev, and M.~Marchal, ``Do vibrotactile patterns on both hands improve guided navigation with a walker?'' in \emph{Proc. EuroHaptics}, ser. Lecture Notes in Computer Science, vol.~14768. Cham, Switzerland: Springer, 2024, pp.~391--404, doi: 10.1007/978-3-031-70058-3\_32.

\bibitem{dupasquier2024scirobot}
C.~du~Pasquier \emph{et al.}, ``Haptiknit: Distributed stiffness knitting for wearable haptics,'' \emph{Sci. Robot.}, vol.~9, no.~97, Art. no. eado3887, 2024, doi: 10.1126/scirobotics.ado3887.

\bibitem{park2025ncomms}
S.~U. Park \emph{et al.}, ``Wearable interactive full-body motion tracking and haptic feedback network systems with deep learning,'' \emph{Nat. Commun.}, vol.~16, Art. no. 8604, 2025, doi: 10.1038/s41467-025-63644-3.

\bibitem{shen2025uist}
V.~Shen and C.~Harrison, ``Kinethreads: Soft full-body haptic exosuit using low-cost motor-pulley mechanisms,'' in \emph{Proc. 38th Annu. ACM Symp. User Interface Softw. Technol. (UIST)}, 2025, Art. no. 1, doi: 10.1145/3746059.3747755.

\bibitem{siviy2023nbme}
C.~Siviy \emph{et al.}, ``Opportunities and challenges in the development of exoskeletons for locomotor assistance,'' \emph{Nat. Biomed. Eng.}, vol.~7, no.~4, pp.~456--472, 2023, doi: 10.1038/s41551-022-00984-1.

\bibitem{molinaro2024nature}
D.~D. Molinaro, K.~L. Scherpereel, E.~B. Schonhaut, G.~Evangelopoulos, M.~K. Shepherd, and A.~J. Young, ``Task-agnostic exoskeleton control via biological joint moment estimation,'' \emph{Nature}, vol.~635, no.~8038, pp.~337--344, 2024, doi: 10.1038/s41586-024-08157-7.

\bibitem{lee2024sciadv}
D.~Lee, S.~Lee, and A.~J. Young, ``AI-driven universal lower-limb exoskeleton system for community ambulation,'' \emph{Sci. Adv.}, vol.~10, no.~51, Art. no. eadq0288, 2024, doi: 10.1126/sciadv.adq0288.

\bibitem{kang2025tro}
I.~Kang \emph{et al.}, ``Online adaptation framework enables personalization of exoskeleton assistance during locomotion in patients affected by stroke,'' \emph{IEEE Trans. Robot.}, vol.~41, pp.~4941--4959, 2025, doi: 10.1109/TRO.2025.3595701.

\bibitem{leestma2024ral}
J.~K. Leestma, S.~Mathur, M.~D. Anderton, G.~S. Sawicki, and A.~J. Young, ``Dynamic duo: Design and validation of an autonomous frontal and sagittal actuating hip exoskeleton for balance modulation during perturbed locomotion,'' \emph{IEEE Robot. Autom. Lett.}, vol.~9, no.~5, pp.~3995--4002, 2024, doi: 10.1109/LRA.2024.3371290.

\bibitem{alili2024tbme}
A.~Alili, A.~Fleming, V.~Nalam, M.~Liu, J.~C. Dean, and H.~Huang, ``Abduction/adduction assistance from powered hip exoskeleton enables modulation of user step width during walking,'' \emph{IEEE Trans. Biomed. Eng.}, vol.~71, no.~1, pp.~334--342, 2024, doi: 10.1109/TBME.2023.3301444.

\bibitem{kucuktabak2024tro}
E.~B. K\"{u}\c{c}\"{u}ktabak \emph{et al.}, ``Haptic transparency and interaction force control for a lower limb exoskeleton,'' \emph{IEEE Trans. Robot.}, vol.~40, pp.~1842--1859, 2024, doi: 10.1109/TRO.2024.3359541.

\bibitem{slade2024nature}
P.~Slade \emph{et al.}, ``On human-in-the-loop optimization of human--robot interaction,'' \emph{Nature}, vol.~633, no.~8031, pp.~779--788, 2024, doi: 10.1038/s41586-024-07697-2.


\bibitem{sarkar2026tmrb}
D.~Sarkar, S.~Chakraborty, S.~Sen, S.~S. Roy, and A.~Arora, ``SoftEXOwalk: A human-in-the-loop optimized soft exosuit for assisting knee flexion using sensorized pneumatic muscles,'' \emph{IEEE Trans. Med. Robot. Bionics}, early access, 2026, doi: 10.1109/TMRB.2026.3722336.

\bibitem{ding2018scirobot}
Y.~Ding, M.~Kim, S.~Kuindersma, and C.~J. Walsh, ``Human-in-the-loop optimization of hip assistance with a soft exosuit during walking,'' \emph{Sci. Robot.}, vol.~3, no.~15, Art. no. eaar5438, 2018, doi: 10.1126/scirobotics.aar5438.

\bibitem{lee2023scirobot}
U.~H. Lee \emph{et al.}, ``User preference optimization for control of ankle exoskeletons using sample efficient active learning,'' \emph{Sci. Robot.}, vol.~8, no.~83, Art. no. eadg3705, 2023, doi: 10.1126/scirobotics.adg3705.

\bibitem{ramella2025icra}
G.~Ramella, A.~Ijspeert, and M.~Bouri, ``Rapid online learning of hip exoskeleton assistance preferences,'' in \emph{Proc. IEEE Int. Conf. Robot. Autom. (ICRA)}, 2025, pp.~2501--2508, doi: 10.1109/ICRA55743.2025.11128809.

\bibitem{wu2024hf}
M.~I. Wu, P.~Stegall, H.~C. Siu, and L.~A. Stirling, ``Impact of haptic cues and an active ankle exoskeleton on gait characteristics,'' \emph{Hum. Factors}, vol.~66, no.~3, pp.~904--915, 2024,doi: 10.1177/00187208221113625.

\bibitem{carrasquillo2025toh}
C.~Carrasquillo, A.~Bajpai, D.~Iyengar, K.~Collins, A.~Mazumdar, and A.~J. Young, ``Enhancing human navigation ability using force-feedback from a lower-limb exoskeleton,'' \emph{IEEE Trans. Haptics}, vol.~18, no.~2, pp.~312--324, 2025, doi: 10.1109/TOH.2025.3533974.

\bibitem{macenski2023ras}
S.~Macenski, T.~Moore, D.~V. Lu, A.~Merzlyakov, and M.~Ferguson, ``From the desks of ROS maintainers: A survey of modern \& capable mobile robotics algorithms in the robot operating system 2,'' \emph{Robot. Auton. Syst.}, vol.~168, Art. no. 104493, 2023, doi: 10.1016/j.robot.2023.104493.

\bibitem{nogueira2014bo}
F.~Nogueira, ``Bayesian Optimization: Open-source constrained global optimization tool for Python,'' GitHub repository, 2014. [Online]. Available: \url{https://github.com/bayesian-optimization/BayesianOptimization}. Accessed: Oct.~19, 2026.
\end{thebibliography}
\end{document}